\documentclass[10pt,conference,letterpaper]{IEEEtran}

\IEEEoverridecommandlockouts
\usepackage{cite}
\usepackage{amsmath,amssymb,amsfonts}
\usepackage{graphicx}
\usepackage{booktabs}
\usepackage{array}
\usepackage{multirow}
\usepackage{url}
\usepackage{xcolor}
\usepackage{balance}
\usepackage{algorithm}
\usepackage{algpseudocode}
\usepackage[hidelinks]{hyperref}

\newcommand{\R}{\mathbb{R}}
\newcommand{\calD}{\mathcal{D}}
\newcommand{\calE}{\mathcal{E}}
\newcommand{\calL}{\mathcal{L}}
\newcommand{\RMSE}{\operatorname{RMSE}}
\newcommand{\MAD}{\operatorname{MAD}}

\begin{document}

\author{
\IEEEauthorblockN{
Khaled Hallak\IEEEauthorrefmark{1},
Oudom Kem\IEEEauthorrefmark{2}
}
\IEEEauthorblockA{
\IEEEauthorrefmark{1}
\textit{American University in Dubai (AUD)}, Dubai, United Arab Emirates\\
Email: khallak@aud.edu
}
\IEEEauthorblockA{
\IEEEauthorrefmark{2}
\textit{CEA LIST}, Palaiseau, France\\
Email: oudom.kem@cea.fr
}

}
\title{Mitigating Catastrophic Forgetting in Task-Free Federated Continual Forecasting via Two-Tier Replay under Concept Drift%
\thanks{This work was supported by the American University in Dubai (AUD).}}

\maketitle

\begin{abstract}
Federated time-series forecasting is increasingly important in applications where temporal data are distributed across privacy-sensitive clients, such as environmental monitoring, smart cities, and industrial sensing. A central limitation of current continual-learning protocols is that they usually assume predefined task boundaries, whereas deployed forecasting systems receive an uninterrupted stream and must decide autonomously when adaptation is necessary. This paper addresses catastrophic forgetting in task-free federated forecasting under concept drift. We introduce a drift-aware continual-learning protocol in which each client monitors prediction-window RMSE, calibrates a robust median threshold from the offline base period, and triggers adaptation only after persistent degradation. To mitigate forgetting, we propose a TwoTierReplay mechanism that separates fixed offline representative memory from online memory. The offline tier preserves long-term knowledge learned during the base phase, while the online tier captures recent distributional changes after drift. Experiments on the Beijing Multi-Site Air-Quality dataset with 12 clients, three forecasting targets, and three horizons show that TwoTierReplay achieves the best stability-plasticity score in all nine target-horizon configurations. It improves over the strongest baseline by approximately 11.4\%. These results suggest that explicit temporal structuring of replay memory is a practical and effective mechanism for task-free federated continual forecasting.
\end{abstract}

\section{Introduction}
Federated learning (FL) enables multiple clients to train predictive models without sharing raw data, making it suitable for distributed sensing, healthcare, smart-city, and environmental monitoring systems \cite{mcmahan2017fedavg}. Many of these applications naturally involve time series: each client observes a local temporal process with its own seasonality, noise level, and drift behavior. Federated time-series forecasting (FTSF) is therefore more difficult than standard supervised FL because the model must cope simultaneously with temporal dependence, non-IID client distributions, and non-stationarity \cite{zhu2021noniid,gama2014conceptdrift}.

Continual learning (CL) provides a suitable framework for adapting forecasting models as new observations arrive. Its main difficulty is catastrophic forgetting, where adaptation to recent data degrades performance on previously learned distributions \cite{besnard2024continualts}. Common catastrophic-forgetting mitigation methods in continual learning include Elastic Weight Consolidation (EWC) \cite{kirkpatrick2017ewc}, Synaptic Intelligence (SI) \cite{zenke2017si}, Learning without Forgetting (LwF) \cite{li2018lwf}, and replay-based strategies \cite{rolnick2019replay}. These methods have been extensively studied in controlled settings; in particular, a previous benchmark studied catastrophic forgetting mitigation in federated time-series forecasting under a controlled task-based protocol, where task boundaries were predefined and several CL methods were compared systematically \cite{hallak2025benchmarking}. However, many task-based benchmarks rely on predefined tasks, known domain shifts, or explicit boundaries. This assumption is often unrealistic for streaming forecasting, where real systems do not announce when the data-generating process has changed.

This paper considers a task-free streaming setting rather than a task-incremental setting. The model observes a continuous stream; no task labels, regime labels, or seasonal boundaries are provided. Adaptation is event-driven: a client updates only when the recent prediction error provides persistent evidence of drift. This setting is more representative of deployed FL forecasting systems, where unnecessary updates can waste computation and increase forgetting, while delayed updates can leave the model misaligned with the current distribution.

Our contribution is a forecasting-specific, drift-aware realization of temporally separated replay rather than the general two-memory principle itself. The replay design separates memory according to its temporal role. A fixed offline memory contains representative base-period samples selected by \(k\)-means and protects long-term knowledge. A bounded online first-in, first-out (FIFO) memory stores recent stream samples and supports rapid adaptation after drift. This separation is important because old and recent observations play different roles under non-stationarity: old samples provide stability, whereas recent samples provide plasticity. We evaluate the framework against Static, Naive fine-tuning, EWC, LwF, SI, and standard single-buffer replay over three forecasting targets and three prediction horizons. The results show that TwoTierReplay consistently improves the stability-plasticity trade-off.

Beyond the proposed algorithm, this study contributes a task-free framework for studying catastrophic forgetting in federated continual time-series forecasting. Unlike task-incremental protocols, the framework does not assume known task or regime boundaries; instead, it evaluates whether a model can adapt from prediction-error evidence while retaining knowledge from the offline base distribution. The evaluation is also designed to test robustness across forecasting difficulty and target dynamics by considering 6-hour, 12-hour, and 24-hour ahead forecasts for three heterogeneous targets: temperature (TEMP), concentration of fine particulate matter with aerodynamic diameter not exceeding 2.5~$\mu$m (PM$_{2.5}$), and wind speed (WSPM). These variables exhibit different temporal characteristics, including smooth seasonality, volatile pollutant behavior, and short-term local variability, making them suitable for assessing the stability-plasticity behavior of continual forecasting methods.

\section{Background, Related Work, and Motivation}
Let $P_{k,t}(\mathbf{X},\mathbf{Y})$ denote the joint distribution
observed by client $k$ at time $t$. Client heterogeneity refers to
variation in $P_{k,t}$ across clients at a fixed time, whereas concept
drift refers to its evolution over time for a fixed client. In
task-free continual learning, the corresponding change points are not
provided to the learner. Plasticity denotes the ability to reduce
prediction error after a change, while stability denotes the ability
to retain performance on previously learned distributions. The
present protocol operationalizes plasticity through online-stream
RMSE and stability through base-period retention.
\subsection{Federated time-series forecasting}
Most FL studies are formulated for classification or static supervised learning. In contrast, FTSF must preserve temporal order and learn from client-specific streams. A monitoring station, household meter, mobile sensor, or industrial device may follow its own seasonal and local dynamics. Directly pooling such data is often impossible because of privacy, ownership, or bandwidth constraints. FL addresses the data-sharing constraint, but it does not by itself solve client heterogeneity or temporal non-stationarity \cite{zhu2021noniid,gama2014conceptdrift}. A model trained during an offline period may become progressively outdated as weather, pollution, sensor behavior, or human activity changes over time. Thus, FTSF requires mechanisms for both distributed training and continual adaptation to address concept drift in local client streams.
\subsection{Recent related work and positioning}

Recent studies have also used multiple learning rates or memory
timescales, although in settings different from the present work.
CLS-ER maintains short- and long-term semantic models that interact
with episodic memory for general, non-federated continual
classification \cite{arani2022clser}. FSNet combines fast adaptation
and associative memory for centralized online time-series forecasting
\cite{pham2023fsnet}. CDA-FedAvg combines client-side concept-drift
detection, short- and long-term memories, and rehearsal for federated
human-activity classification \cite{casado2022cdafedavg}. In contrast,
TwoTierReplay addresses multi-step federated forecasting: persistent
forecast-error degradation triggers adaptation, a fixed sample-level
offline tier protects the base distribution, and a bounded online FIFO
tier represents recent stream behavior. Accordingly, the contribution
claimed here is the integration and evaluation of this memory
organization within a causal, task-free federated forecasting
protocol, rather than the first use of dual-timescale memory.
\subsection{Catastrophic forgetting in continual learning}
Continual learning provides a mechanism for adapting forecasting models to concept drift as new observations arrive. However, sequential adaptation introduces the stability-plasticity dilemma. A highly plastic model adapts quickly to new observations but may overwrite previously useful parameters; a highly stable model retains old knowledge but may fail under drift \cite{besnard2024continualts}. EWC and SI reduce forgetting by penalizing changes to important parameters \cite{kirkpatrick2017ewc,zenke2017si}. LwF regularizes the adapted model using predictions from a previous teacher \cite{li2018lwf}. Replay stores or generates representative samples from previous distributions and mixes them with new data during training \cite{rolnick2019replay}. In task-incremental benchmarks, these approaches are usually evaluated after each known task. That evaluation is valuable for controlled comparison, but it does not directly match streaming forecasting, where task boundaries are not available.

\subsection{Why task-free evaluation is necessary}
Task-based protocols provide adaptation times, whereas deployed forecasters receive neither task labels nor drift boundaries \cite{aljundi2019taskfree}. Our protocol therefore infers when to update from persistent prediction-error degradation and evaluates whether the final model preserves the offline base distribution while maintaining low online-stream error.

\section{Task-Free Federated Forecasting Problem}
Consider $K$ clients. Client $k$ observes a local multivariate time series
\begin{equation}
    \calD_k=\{(\mathbf{x}_{t}^{k},y_{t}^{k})\}_{t=1}^{T_k}, \quad
    \mathbf{x}_{t}^{k}\in\R^d,\; y_t^k\in\R .
\end{equation}
A forecasting model $f_{\theta}$ maps a lagged input window to a future target sequence. For an input length $n$ and prediction horizon $p$, samples are constructed as
\begin{equation}
    \mathbf{X}_{t}^{k}=[\mathbf{x}_{t-n}^{k},\ldots,\mathbf{x}_{t-1}^{k}],\quad
    \mathbf{Y}_{t}^{k}=[y_t^k,\ldots,y_{t+p-1}^k].
\end{equation}
The online goal is to update $\theta$ as data arrive while minimizing forecasting loss across clients,
\begin{equation}
    \min_{\theta_t}\sum_{k=1}^{K} \calL_k(f_{\theta_t};\calD_k^{\le t}),
\end{equation}
subject to three constraints: clients cannot retain the full historical stream, no task boundaries are known, and the distribution may evolve over time.

At a communication event, client $k$ receives the current global model, performs local optimization, and returns parameters $\theta_k$. The server aggregates them by weighted FedAvg \cite{mcmahan2017fedavg},
\begin{equation}
    \theta^{+}=\sum_{k=1}^{K} \frac{n_k}{\sum_{j=1}^{K}n_j}\theta_k,
\end{equation}
where $n_k$ is the number of local samples used in the update. In the task-free protocol, this aggregation is not performed after every stream window; it is performed only when the drift detector confirms that adaptation is needed. Thus, the method combines the privacy-preserving optimization structure of FL with an event-driven continual-learning schedule.

\begin{figure*}[!t]
\centering
\includegraphics[width=0.82\textwidth]{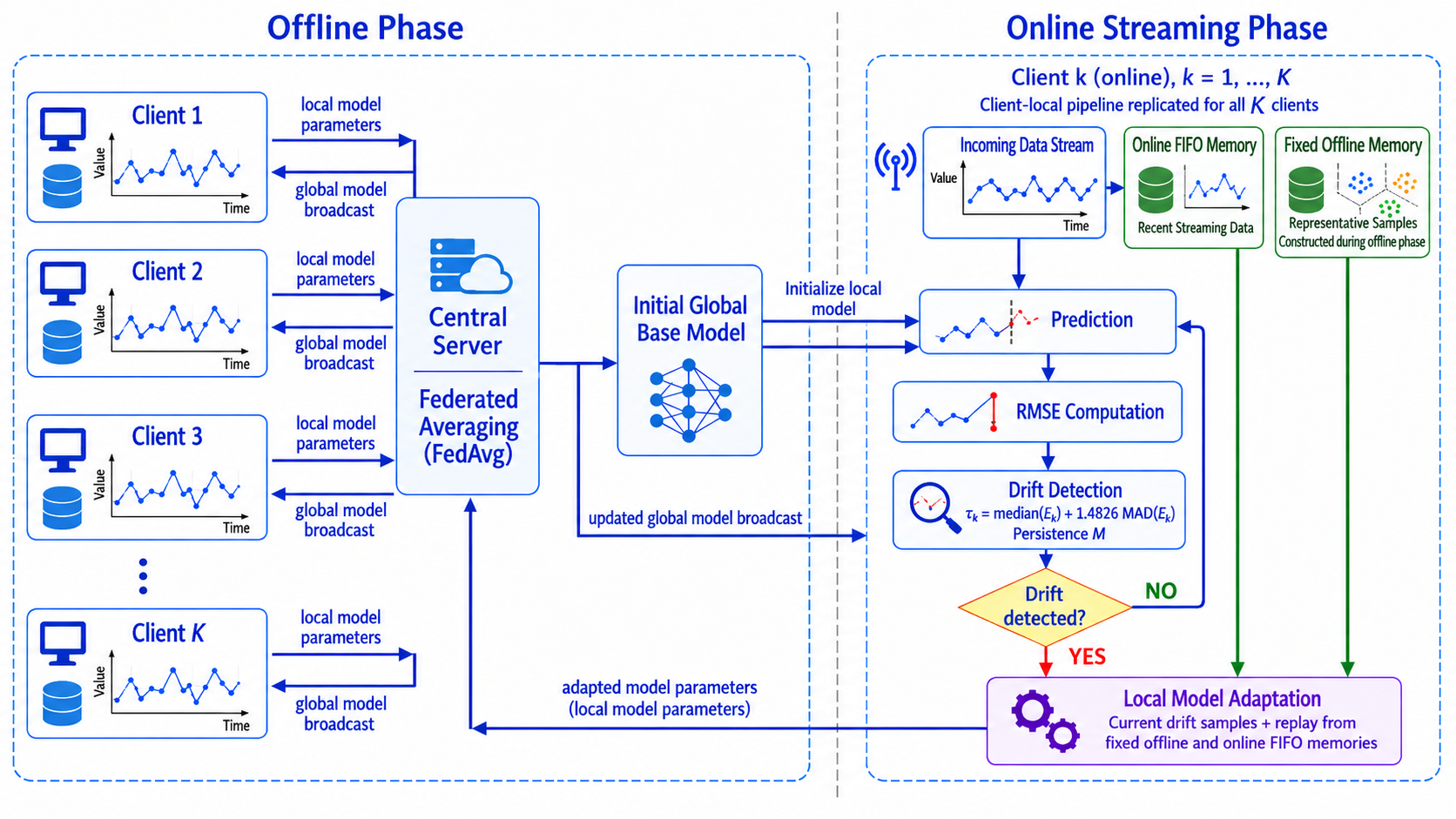}
\caption{Task-free federated continual-forecasting protocol. Each client maintains fixed offline and bounded online replay memories locally. A persistent drift alarm at any client triggers local adaptation at all $K$ clients; the server aggregates only model parameters and rebroadcasts the updated global model.}
\label{fig:framework}
\end{figure*}

Concept drift \cite{gama2014conceptdrift} is inferred from prediction error because the true data-generating distribution is not observable. At stream index $t$, the prediction-window RMSE for client $k$ is
\begin{equation}
    e_{k,t}=
    \sqrt{\frac{1}{p}\sum_{h=0}^{p-1}
    \left(y_{k,t+h}-\hat{y}_{k,t+h}\right)^2}.
\end{equation}

A persistent increase in $e_{k,t}$ indicates that the current model may no longer match the local stream distribution. This error-based signal is used both for plasticity evaluation and for triggering adaptation.
\section{Proposed Drift-Aware Two-Tier Replay}
\subsection{Overall offline-online framework}
The proposed framework is organized into two coupled phases, as illustrated in Fig.~\ref{fig:framework}. In the offline phase, the clients use the base-period data to train an initial global forecasting model through FedAvg. This model represents the knowledge available before deployment and is later used as the reference model for drift calibration and forgetting evaluation.

During the same offline phase, each client builds a compact memory from its base-period data. Instead of storing the full historical sequence, a small set of representative forecasting samples is selected using the \(k\)-means-based sampling strategy adopted from \cite{hallak2025ppbc}. The resulting offline memory contains representative input-output pairs from the base distribution. This memory remains fixed throughout the online phase and is used to preserve long-term knowledge during later CL adaptation.

During the online streaming phase, the deployed global model receives new windows chronologically and forecasts the next $p$ values, where $p$ is the selected prediction horizon. These forecasts are made before the corresponding ground-truth values are available. Once the true values are observed, each client computes the prediction-window RMSE and compares it with its calibrated drift threshold. If the drift is not persistent, the model continues forecasting without update. If persistent drift is detected, continual adaptation is triggered using the recent stream samples together with the available replay mechanism.

\subsection{Robust drift detection}
After offline training, each client calibrates its own drift threshold using the prediction errors obtained on its base-training data.
For client $k$, let
\begin{equation}
\calE_k=\{e_{k,1},e_{k,2},\ldots,e_{k,N_k}\},
\end{equation}
denote the set of RMSE values computed over all valid prediction windows in the base-training period. The median calibration error is
\begin{equation}
m_k=\operatorname{median}(\calE_k),
\end{equation}
and the median absolute deviation is
\begin{equation}
\MAD(\calE_k)
=
\operatorname{median}
\left(
|e_{k,1}-m_k|,\ldots,|e_{k,N_k}-m_k|
\right).
\end{equation}
The client-specific drift threshold is then
\begin{equation}
\tau_k=m_k+1.4826\,\MAD(\calE_k),
\label{eq:tau}
\end{equation}
where $\tau_k$ is the local error threshold for client $k$. The constant 1.4826 is the standard normal consistency factor that scales median absolute deviation (MAD) to be comparable to the standard deviation under approximately Gaussian calibration errors \cite{rousseeuw1993alternatives}.

During online streaming, $e_{k,t}$ denotes the prediction-window RMSE computed at stream time $t$ after the true target values become available. A drift alarm is raised only if
\begin{equation}
e_{k,t}>\tau_k
\quad \text{for } M \text{ consecutive prediction windows}.
\end{equation}
Here, $M$ is the persistence parameter. It prevents isolated error spikes or outliers from triggering unnecessary adaptation and ensures that updates are performed only after sustained degradation in forecasting performance.

While drift detection determines \emph{when} to adapt, it does not address \emph{what} knowledge to preserve during adaptation. To mitigate forgetting, we augment drift-triggered updates with a two-tier replay mechanism, described next, which separates long-term base knowledge from recent stream information.

\subsection{Two-tier replay memory}
\label{sec:twotier}
The proposed replay memory contains two components. The offline tier is fixed after initialization and stores representative base-period samples selected during the offline phase. This tier preserves long-term knowledge learned before deployment. The online tier is a bounded FIFO queue that stores recent stream samples and is updated as new observations arrive.
When drift is confirmed, adaptation is performed using the recent stream samples associated with the persistent RMSE increase. The loss computed on these samples is denoted by $\calL_{\rm cur}$. In the Naive continual-learning baseline, the model is updated using only $\calL_{\rm cur}$, meaning that adaptation depends only on the current drift-related stream data and has no explicit mechanism to preserve base-period knowledge.

\begin{figure}[!t]
\centering
\includegraphics[width=\columnwidth]{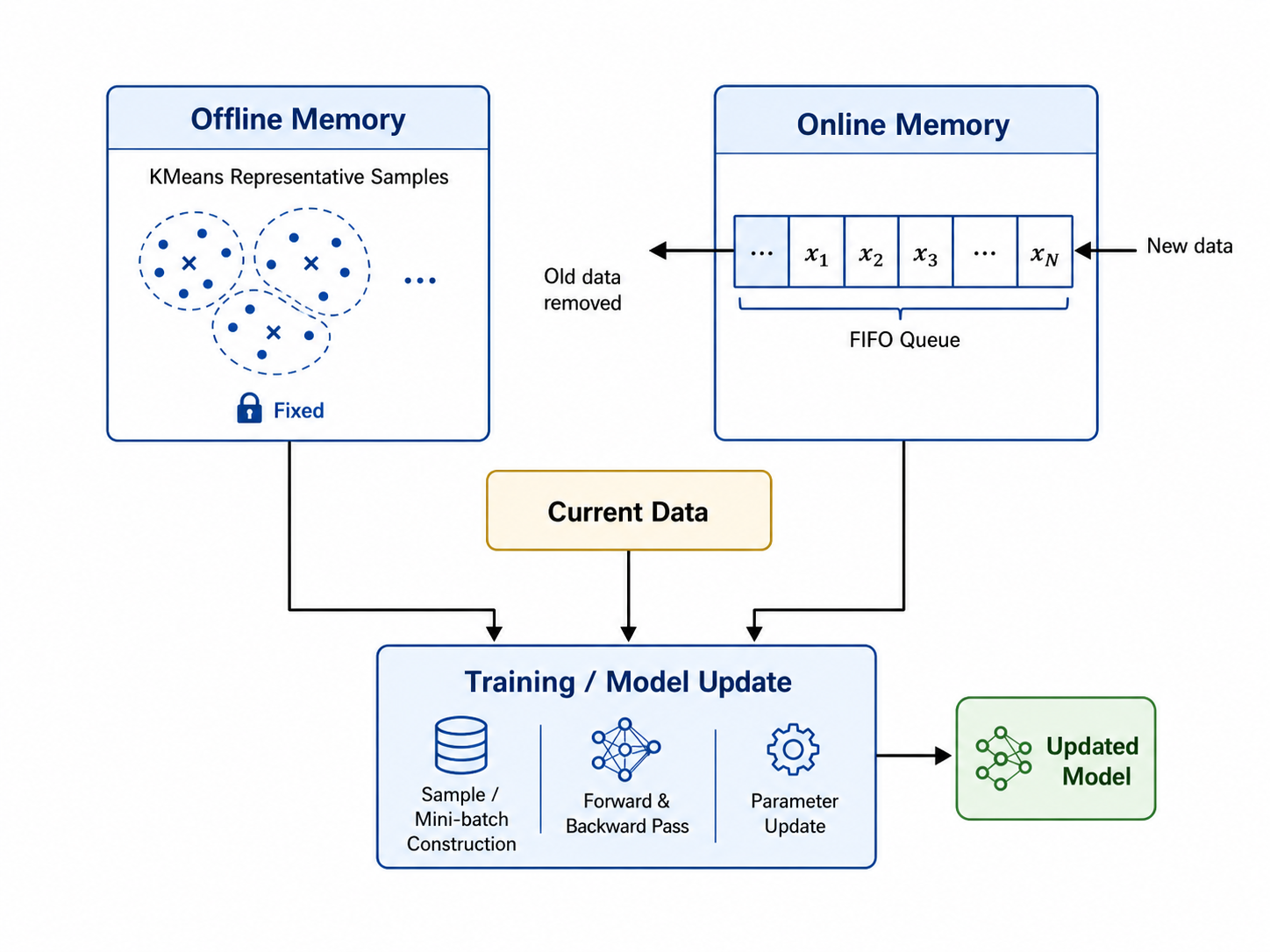}
\caption{TwoTierReplay memory and local-update flow. The non-evictable offline tier stores representative base-period samples, whereas the bounded FIFO tier stores recently revealed stream samples. At a drift event, current samples and replay samples from both tiers are used for local adaptation.}
\label{fig:twotier}
\end{figure}

Replay-based methods extend this update by adding a replay loss $\calL_{\rm rep}$, computed from stored samples \cite{rolnick2019replay}. In TwoTierReplay, each replay mini-batch is constructed by sampling from two memory tiers. For each replay sample, the offline tier is selected with probability $p_{\rm off}$, while the online FIFO tier is selected with probability $1-p_{\rm off}$. Thus, $p_{\rm off}$ controls the balance between long-term retention and recent adaptation. In our experiments, $p_{\rm off}=0.8$, meaning that replay is biased toward representative base-period samples while still including recent online samples when available. Once both tiers are nonempty, $p_{\rm off}=0.8$ gives an expected
offline-to-online replay-sample ratio of $4{:}1$. This fixed setting
deliberately biases adaptation toward retention of base-period
knowledge while reserving 20\% of replay draws for recent stream
behavior. It was kept unchanged across targets and horizons and is
not claimed to be universally optimal. The adaptation objective is
\begin{equation}
\calL
=
\lambda_{\rm cur}\calL_{\rm cur}
+
\lambda_{\rm rep}\calL_{\rm rep}.
\label{eq:loss}
\end{equation}
The weights $\lambda_{\rm cur}\geq 0$ and
$\lambda_{\rm rep}\geq 0$ control the contributions of the current
and replay losses; they are selected according to the procedure in
Section~VI-D and are not required to sum to one. 
This design combines current drift-related data with long-term and recent replay. NormalReplay stores base and stream samples in one fixed-capacity FIFO buffer, so base samples are eventually evicted. TwoTierReplay instead keeps the \(k\)-means-selected offline tier non-evictable and limits the online tier to the 300 most recent samples (Fig.~\ref{fig:twotier}).

\subsection{Event-driven federated adaptation}
When at least one client confirms drift, the server initiates a federated adaptation round. The drift alarm is used as a communication trigger, not as an assumption that all clients have drifted. The current global model is broadcast to all $K$ clients. Drifted clients adapt using the recent stream samples associated with the persistent RMSE increase, while non-drifted clients participate using their recent local samples and the method-specific retention mechanism.

This design keeps the update globally balanced. If only drifted clients were aggregated, the global model could move too strongly toward a local distributional change and harm clients whose data remain stable. Including non-drifted clients provides a stabilizing effect in FedAvg: drifted clients contribute plasticity, whereas non-drifted clients help preserve global stability.

Each client performs local continual adaptation using the selected method. For replay-based methods, the client trains with current stream samples and replay samples. For EWC and SI, the current loss is augmented with the corresponding parameter-importance penalty. For LwF, the current model is regularized toward teacher predictions from the previous model. The server then aggregates the updated client models using FedAvg. If no client detects drift, no communication or training step is performed and the current model continues forecasting. The complete task-free procedure is summarized in Algorithm~\ref{alg:taskfree}.
\begin{algorithm}[t]
\caption{Task-free drift-aware federated continual forecasting}
\label{alg:taskfree}
\footnotesize
\begin{algorithmic}[1]
\Statex \textbf{Input:} Base data, stream data, clients $1,\ldots,K$
\Statex \textbf{Input:} Prediction horizon $p$ and persistence parameter $M$
\Statex \textbf{Output:} Final adapted global model and evaluation metrics

\State Train the offline global model using FedAvg.
\For{each client $k$}
\State Compute base-period calibration RMSE values.
\State Compute the threshold $\tau_k$ using \eqref{eq:tau}.
\State Build the offline memory from representative samples.
\EndFor

\For{each online stream window}
\For{each client $k$}
\State Forecast the next $p$ values.
\State Compute $e_{k,t}$ when ground truth is available.
\State Append the newly revealed forecasting sample to client $k$'s
online FIFO and evict the oldest entry if its capacity exceeds 300.
\State Update the drift counter for client $k$.
\EndFor

\If{at least one client satisfies the persistence rule}
    \State Broadcast the current global model to all $K$ clients.
    \State Each client performs the method-specific local update.
    \State Aggregate all $K$ updated client models using FedAvg.
\Else
    \State Continue forecasting without model update.
\EndIf

\EndFor

\State Evaluate stream RMSE and base-retention forgetting.
\end{algorithmic}
\end{algorithm}

\section{Continual-Learning Strategies}
The proposed framework provides a common task-free protocol for evaluating and comparing the main principles of catastrophic-forgetting mitigation in continual learning. The selected baselines cover the main forgetting-mitigation methods considered in the federated time-series forecasting benchmark \cite{hallak2025benchmarking}.

\subsection{Naive adaptation}
Naive adaptation updates the model using only the recent stream samples collected around the detected drift. Its local loss is
\begin{equation}
\calL_{\rm naive}=\calL_{\rm cur}.
\end{equation}
This method has high plasticity but no explicit mechanism to protect previously learned knowledge. It is therefore a useful reference baseline for forgetting-mitigation methods.

\subsection{Regularization-based methods}
EWC penalizes deviations from parameters that were important for the base model \cite{kirkpatrick2017ewc}. Its adaptation loss is
\begin{equation}
\calL_{\rm EWC}=\calL_{\rm cur}+\lambda_{\rm EWC}
\sum_i F_i(\theta_i-\theta_i^{0})^2 ,
\end{equation}
where $F_i$ is a diagonal Fisher estimate and $\theta^0$ is the reference base parameter vector. SI has a similar role but uses path-integral importance weights $\Omega_i$ accumulated during training \cite{zenke2017si}:
\begin{equation}
\calL_{\rm SI}=\calL_{\rm cur}+\lambda_{\rm SI}
\sum_i \Omega_i(\theta_i-\theta_i^{0})^2.
\end{equation}
Both methods can reduce destructive parameter drift, but neither reintroduces explicit historical samples. In non-stationary forecasting, this may be insufficient when the input-output patterns of the base period must be rehearsed.

\subsection{Distillation-based method}
LwF uses the previous model as a teacher and penalizes disagreement between teacher and student predictions on recent samples \cite{li2018lwf}:
\begin{equation}
    \calL_{\rm LwF}
    =
    \calL_{\rm cur}
    +
    \lambda_{\rm LwF}
    \|f_{\theta}(X)-f_{\theta^{-}}(X)\|_2^2 ,
\end{equation}
where \(\theta^{-}\) denotes the model parameters before the current drift-triggered adaptation event.

This avoids storing raw historical data, but its effectiveness depends on the relevance of the teacher. Under drift, a teacher trained on an earlier distribution may not always provide useful targets for the current stream.

\subsection{Replay-based methods}
Replay-based methods reduce forgetting by rehearsing stored samples together with current data.
The NormalReplay baseline used in this study \cite{rolnick2019replay} uses a single replay buffer constructed from representative base-period samples. Its loss has the same form as (\ref{eq:loss}), but replay examples are drawn from one undifferentiated memory. TwoTierReplay differs by separating the fixed offline memory from the online FIFO memory. This prevents representative base samples from being overwritten by recent stream samples while still allowing recent drift-related samples to contribute to adaptation. This is the main methodological distinction tested in the experiments.

\section{Experimental Setup}
\subsection{Dataset and preprocessing}
Experiments use the Beijing Multi-Site Air-Quality dataset from the UCI Machine Learning Repository \cite{chen2017beijing}. The dataset contains hourly air-pollutant and meteorological measurements from 12 nationally controlled monitoring stations between March 2013 and February 2017. Each station is treated as one federated client. The forecasting targets are temperature (TEMP), fine particulate matter concentration (PM$_{2.5}$), and wind speed (WSPM), which represent different temporal behaviors: smoother seasonal meteorological dynamics, volatile pollutant dynamics, and locally variable wind dynamics.

Missing offline numerical values are imputed using time-aware interpolation, with forward/backward filling when needed. Wind direction is converted into sine and cosine components. Hour, month, and day variables are cyclically encoded. All features are scaled using robust percentile min-max bounds based on the first and 99th percentiles computed only from the offline base period. These scaling bounds are then fixed and applied to each online observation when it arrives; the scaler is not refitted during streaming, so no information from future online observations is used during prediction or adaptation.

\subsection{Streaming protocol and model}
The data are processed as an hourly chronological stream. The offline training period covers 1 March 2013 to 1 March 2014, while the online stream covers 2 March 2014 to 28 February 2017. Each input window contains the previous $n=12$ hourly observations, and the prediction horizons are $p\in\{6,12,24\}$, corresponding to 6-hour, 12-hour, and 24-hour ahead forecasting. These horizons allow us to evaluate whether the stability-plasticity behavior remains consistent as the forecasting task becomes more difficult, from short-term to longer look-ahead prediction.

The proposed framework is not tied to a specific model architecture. Any model that can be trained locally and aggregated in a federated setting can be used as the forecasting model. In this study, we use a one-layer long short-term memory (LSTM) with hidden dimension 64 and a linear output head as a standard recurrent baseline for multivariate time-series forecasting. This choice allows the experiments to focus on the effect of the continual-learning strategy rather than on architectural complexity.

During the offline phase, the base model is first trained using FedAvg for 100 communication rounds with one local epoch per round and learning rate $10^{-4}$, before any continual-learning adaptation is performed. Continual adaptation uses five communication rounds, one local epoch, and learning rate $10^{-3}$. Experiments are repeated over three random seeds to reduce the effect of random initialization, stochastic optimization, and memory selection. All methods use the same seeds and streaming protocol to ensure a fair comparison.

The memory sizes refer to the number of forecasting samples stored in each replay buffer. For the offline tier, representative samples are selected from 10\% of the base-period data. Tuning indicated that using approximately 10--20\% of the offline data provides a compact but representative memory, while avoiding unnecessary storage of redundant historical windows. We select 10\% as the default to keep the replay memory efficient. For the online tier, the FIFO capacity is fixed at 300 recent stream samples, which provided the best balance between retaining recent drift-related information and avoiding excessive dependence on older stream samples. The offline replay sampling probability is set to $p_{\rm off}=0.8$ so that replay favors long-term base knowledge while still incorporating recent online information.

All clients use the same model architecture and preprocessing pipeline. This avoids confounding the comparison with client-specific model choices and allows the observed differences to be attributed to the continual-learning strategy. The base period is used only for offline training, calibration, and retention evaluation. The online stream is processed chronologically and is never shuffled. This preserves the causal structure of deployment: predictions are made before the corresponding targets are revealed, and adaptation is allowed only after ground truth data becomes available. Table~\ref{tab:config} summarizes the shared experimental configuration.

\begin{table}[t]
\centering
\caption{Main experimental configuration.}
\label{tab:config}
\scriptsize
\begin{tabular}{ll}
\toprule
Component & Setting \\
\midrule
Clients & 12 monitoring stations \\
Targets & TEMP, PM$_{2.5}$, WSPM \\
Horizons & $p=6,12,24$ hours \\
Input length & $n=12$ hours \\
Model & LSTM, hidden dimension 64 \\
Base training & FedAvg, 100 rounds, 1 local epoch \\
Base learning rate & $10^{-4}$ \\
Continual adaptation & 5 rounds, 1 local epoch \\
CL learning rate & $10^{-3}$ \\
Drift threshold & median + $1.4826\,\MAD$ \\
Persistence & $M=2$ Streams \\
Base sample fraction & 0.10 \\
Online FIFO capacity & 300 samples \\
Offline replay probability & $p_{\rm off}=0.8$ \\
Random seeds & 42, 43, 44 \\
Optimizer & Adam\\
Batch size & 32\\
\bottomrule
\end{tabular}
\end{table}

\subsection{Baselines and metrics}
We compare seven methods. Static evaluates the offline model without adaptation. Naive CL updates only on recent stream samples. EWC and SI represent parameter-importance regularization \cite{kirkpatrick2017ewc,zenke2017si}, LwF represents teacher-based distillation \cite{li2018lwf}, and NormalReplay represents standard experience replay \cite{rolnick2019replay}. TwoTierReplay uses the proposed fixed-offline plus online-FIFO structure. This comparison tests whether replay is beneficial and whether separating long-term and recent memory improves the stability--plasticity balance.

Because no task boundaries exist, task-incremental metrics such as average performance and average forgetting are not appropriate \cite{aljundi2019taskfree}. We therefore report stream RMSE for plasticity and base-retention forgetting for stability:
\begin{equation}
    AF_{\rm base}=\RMSE_{\rm base}^{f}-\RMSE_{\rm base}^{0},
\end{equation}
where $\RMSE_{\rm base}^{0}$ is the offline model's RMSE on the base distribution and $\RMSE_{\rm base}^{f}$ is the final adapted model's RMSE on the same base distribution. The main metric is the signed stability-plasticity score
\begin{equation}
    SP=\RMSE_{\rm stream}+ AF_{\rm base}
\end{equation}

Lower $SP$ indicates a better balance between online adaptation and base-period retention. Positive $AF_{\mathrm{base}}$ indicates forgetting on the base distribution, whereas negative $AF_{\mathrm{base}}$ indicates that the final adapted model improves on the base-period evaluation.

\subsection{Hyperparameter tuning}
To ensure a fair comparison, all continual-learning methods were tuned under the same task-free streaming protocol. Method-specific hyperparameters, including the current-data weight $\lambda_{\rm cur}$, replay weight $\lambda_{\rm rep}$, EWC regularization weight $\lambda_{\rm EWC}$, LwF distillation weight $\lambda_{\rm LwF}$, SI regularization weight $\lambda_{\rm SI}$, replay fraction, and offline replay probability $p_{\rm off}$, were selected empirically using the same validation procedure across methods. Since the best configurations vary across forecasting targets and prediction horizons, a full hyperparameter table is omitted from the main paper due to space limitations. To support reproducibility, the accompanying GitHub repository \footnote{\url{https://github.com/khaledhallakk/TaskFree-Federated-Continual-Forecasting}} provides the complete notebook implementation, preprocessing pipeline, hyperparameter configuration, random-seed protocol, supplementary experimental details, and the detailed result files. 

\section{Results and Discussion}
Table~\ref{tab:sp} reports the mean signed stability--plasticity score over three seeds. TwoTierReplay achieves the lowest score in all nine target--horizon configurations and the lowest overall average (0.1559), an 11.4\% reduction relative to NormalReplay, the strongest baseline (0.1760). This consistent margin shows that the improvement is not merely due to replay, but to the organization of replay memory according to temporal function. NormalReplay confirms that replay is useful; however, as described in Section~\ref{sec:twotier}, its base-period samples are progressively evicted as stream samples accumulate, weakening stability over time. TwoTierReplay avoids this by permanently protecting the offline tier while bounding the online tier to recent observations, which explains the consistent improvement across all configurations. Therefore, the model continues to rehearse stable representative base samples while also adapting to recent drift-related samples. 

\begin{table*}[t]
\centering
\caption{Mean signed stability--plasticity score, $SP=\RMSE_{\rm stream}+AF_{\rm base}$, over three random seeds. The Avg. column is the arithmetic mean across the nine target-horizon configurations. Lower values are better.}
\label{tab:sp}
\scriptsize
\setlength{\tabcolsep}{3.2pt}
\begin{tabular}{lcccccccccc}
\toprule
Method & TEMP-6 & TEMP-12 & TEMP-24 & PM$_{2.5}$-6 & PM$_{2.5}$-12 & PM$_{2.5}$-24 & WSPM-6 & WSPM-12 & WSPM-24 & Avg. \\
\midrule
Static       & 0.2413 & 0.2430 & 0.2503 & 0.1582 & 0.1620 & 0.1660 & 0.1504 & 0.1569 & 0.1637 & 0.1880 \\
Naive        & 0.2139 & 0.2292 & 0.2432 & 0.1614 & 0.1642 & 0.1686 & 0.1767 & 0.1678 & 0.1723 & 0.1886 \\
EWC          & 0.2043 & 0.2285 & 0.2425 & 0.1532 & 0.1561 & 0.1619 & 0.1512 & 0.1580 & 0.1638 & 0.1799 \\
LwF          & 0.2139 & 0.2292 & 0.2434 & 0.1567 & 0.1606 & 0.1631 & 0.1742 & 0.1741 & 0.1801 & 0.1884 \\
SI           & 0.2139 & 0.2293 & 0.2430 & 0.1588 & 0.1605 & 0.1639 & 0.1553 & 0.1580 & 0.1632 & 0.1829 \\
NormalReplay & 0.1966 & 0.2099 & 0.2186 & 0.1502 & 0.1562 & 0.1637 & 0.1577 & 0.1627 & 0.1680 & 0.1760 \\
TwoTierReplay& \textbf{0.1827} & \textbf{0.1939} & \textbf{0.2006} & \textbf{0.1083} & \textbf{0.1352} & \textbf{0.1515} & \textbf{0.1316} & \textbf{0.1448} & \textbf{0.1546} & \textbf{0.1559} \\
\bottomrule
\end{tabular}
\end{table*}

The advantage of TwoTierReplay is strongest for PM$_{2.5}$, which exhibits more volatile and less seasonal dynamics. Nevertheless, it also achieves the lowest SP scores for TEMP and WSPM across all horizons, indicating that the benefit of memory separation is not limited to highly variable targets. The fixed offline tier preserves representative base-period patterns, whereas the bounded online FIFO emphasizes recent stream behavior. The consistent improvement over NormalReplay suggests that separating these temporal roles provides a better stability--plasticity balance across targets with different dynamics, although event-level analysis would be required to attribute the magnitude of improvement to specific pollution or meteorological phenomena.

The comparison with Static is also informative. Static is competitive for some WSPM configurations, showing that adaptation is not always beneficial when the offline model already generalizes reasonably well. However, Static cannot respond to persistent distributional changes and is clearly inferior on TEMP and PM$_{2.5}$. Naive fine-tuning improves TEMP compared with Static, but it harms WSPM and does not consistently improve PM$_{2.5}$. This behavior is typical of uncontrolled plasticity: the model can follow recent samples but lacks a mechanism to protect old knowledge or stabilize adaptation.

As shown in Table~\ref{tab:sp}, EWC and SI perform moderately well but have higher average SP scores than the replay-based methods. This may be because EWC and SI protect important parameters, whereas replay methods explicitly reintroduce representative historical samples during adaptation \cite{kirkpatrick2017ewc,zenke2017si,rolnick2019replay}. In this non-stationary forecasting setting, our results suggest that parameter protection alone may be less effective than directly rehearsing earlier input-output patterns. LwF is also limited because the teacher model is inherited from an earlier stream state; if the teacher is misaligned with a new local regime, distillation may preserve outdated predictions \cite{li2018lwf}. These observations help explain why replay-based approaches are more effective in the present controlled experiments.

Horizon length also affects performance. For TwoTierReplay, the score increases from 0.1827 to 0.2006 for TEMP, from 0.1083 to 0.1515 for PM$_{2.5}$, and from 0.1316 to 0.1546 for WSPM when moving from horizon 6 to horizon 24. This is expected because longer horizons introduce greater uncertainty and error accumulation. Nevertheless, TwoTierReplay remains the best method at all horizons, suggesting that the replay structure is robust to increasing forecasting difficulty.
\subsection{Practical implications}
The proposed framework is applicable to real monitoring networks because it converts persistent forecasting-error degradation into an operational trigger for model updates, removing the need for predefined seasonal tasks or manually labeled drift periods. In federated settings, this event-driven schedule can avoid unnecessary communication and local retraining. The two-tier memory also has a clear functional interpretation: the fixed offline tier preserves representative base-period knowledge, whereas the bounded FIFO tier retains recent, replaceable stream samples. Unlike a single buffer, this separation prevents recent observations from progressively displacing representative base samples, which is consistent with the observed improvement over NormalReplay.

Beyond the main stability--plasticity score, the implementation records stream RMSE, base-retention forgetting, detected drifts, global adaptation events, and CPU time. These diagnostics help distinguish predictive improvement from excessive retraining, while the main paper focuses on the stability--plasticity score because it directly summarizes the central objective of adapting to the online stream while retaining base-period knowledge.

\subsection{Limitations}
The present study focuses on one multivariate environmental dataset and one forecasting model. Although the dataset contains 12 heterogeneous clients and multiple targets, additional datasets are needed to confirm generality across domains such as energy demand, traffic, finance, and industrial sensor monitoring. The framework is also evaluated using an LSTM model; future studies should examine whether similar stability-plasticity behavior is obtained with other forecasting architectures, such as temporal convolutional networks, Transformers, or lightweight recurrent models.

Another limitation concerns computational cost. Although TwoTierReplay achieves the best stability-plasticity performance among the evaluated methods, its replay-based adaptation requires additional memory sampling and training computations compared with simpler baselines such as Naive adaptation, EWC, SI, or LwF. This is especially relevant in task-free federated settings, where drift-triggered updates may occur at different clients and communication or local computation can become costly. Promising directions include reducing replay frequency, adaptive replay scheduling, latent replay, prototype-based memories, and lightweight generative surrogates. Such approaches may reduce the dependence on explicit sample replay while preserving the benefits of historical knowledge retention.

The threshold in (\ref{eq:tau}) is robust and simple, but adaptive thresholding could improve sensitivity when the base calibration distribution is not representative of future stream variability. Finally, this study uses a shared global aggregation protocol; future work should investigate personalized aggregation and client-specific drift handling.

\section{Conclusion}
This paper presented a task-free federated continual-learning framework for time-series forecasting under concept drift. Instead of assuming predefined tasks, the framework monitors prediction-window RMSE and triggers adaptation only after persistent robust-threshold violations. The proposed TwoTierReplay mechanism separates long-term offline representative memory from short-term online FIFO memory, allowing the model to preserve base-period knowledge while adapting to recent distributional changes. Experiments on 12 Beijing air-quality clients, three forecasting targets, and three horizons show that TwoTierReplay achieves the best stability-plasticity score in every evaluated configuration, with an 11.4\% average improvement over standard replay. These findings suggest that drift-aware, temporally structured replay is a promising direction for deployable federated forecasting systems. Future work will extend the framework to additional datasets, adaptive threshold selection, communication-efficient replay sampling, and personalized client-level adaptation.

\balance
\bibliographystyle{IEEEtran}
\bibliography{references}

\end{document}